\documentclass[conference]{IEEEtran}

\ifCLASSINFOpdf
  \usepackage[pdftex]{graphicx}
  \DeclareGraphicsExtensions{.pdf,.jpeg,.png}
\else
\fi

\usepackage{amsfonts, graphicx,amsmath,amsbsy,amssymb}

\newcommand{\eb}{{\mathbf e}}

\newcommand{\xb}{{\mathbf x}}
\newcommand{\yb}{{\mathbf y}}

\newcommand{\Mc}{{\mathcal M}}
\newcommand{\Nc}{{\mathcal N}}

\newcommand{\Qc}{{\mathcal Q}}
\newcommand{\Rc}{{\mathcal R}}

\newcommand{\Tc}{{\mathcal T}}

\newcommand{\Wc}{{\mathcal W}}

\newcommand{\thetab}{{\boldsymbol{\theta}}}

\newcommand{\Rd}{{\mathbb R}}

\newcommand{\Sd}{{\mathbb S}}

\begin{document}
%
\title{Deep Learning   Interior Tomography for Region-of-Interest Reconstruction}

\author{
\IEEEauthorblockN{Yoseob Han}
\IEEEauthorblockA{
KAIST, Daejeon, Korea\\
Email: hanyoseob@kaist.ac.kr}
\and
\IEEEauthorblockN{Jawook Gu}
\IEEEauthorblockA{
KAIST,  Daejeon, Korea\\
Email: jwisdom9299@kaist.ac.kr}
\and
\IEEEauthorblockN{Jong Chul Ye}
\IEEEauthorblockA{
KAIST, Daejeon, Korea\\
Email: jong.ye@kaist.ac.kr}}



\maketitle

\begin{abstract}
Interior tomography for the region-of-interest imaging has advantages of using a small detector and reducing X-ray radiation dose.
However, standard analytic reconstruction suffers from severe cupping artifacts due to existence of null space in the
truncated Radon transform.
Existing penalized reconstruction methods may address this problem but they require extensive computations due to the iterative reconstruction.
Inspired by the recent deep learning approaches  to low-dose and sparse view CT,
here we propose a deep learning architecture 
that removes  null space signals from the FBP reconstruction.
Experimental results have shown that the proposed method provides near-perfect reconstruction with about $7\sim 10$dB improvement in PSNR over existing methods
in spite of significantly reduced
run-time complexity.
\end{abstract}


%
\IEEEpeerreviewmaketitle

\section{Introduction}
X-ray Computed Tomography (CT) is one of the most powerful clinical imaging imaging tools, delivering high-quality images in a fast and cost effective manner.
However,  the X-ray is harmful to the human body, so many studies has been conducted to develop methods that reduce the X-ray dose.
Specifically, X-ray doses can be reduced by reducing the number of photons, projection views  or the size of the field-of-view of X-rays. 
Among these, the CT technique for reducing the field-of-view of X-ray is called interior tomography. 
Interior tomography is useful when the region-of-interest (ROI) within a patient's body is small (such as heart), because
interior tomography aims to obtain an ROI image by irradiating only the ROI with x-rays.
Interior tomography not only can dramatically reduce the X-ray dose, 
but also has cost benefits by using a small-sized detector.
However, 
the use of an analytic CT reconstruction algorithm generally produces images with severe artifacts due to the transverse directional
projection truncation.


Sinogram extrapolation is a simple approximation method to reduce the artifacts. 
However, sinogram extrapolation method still generates biased CT number in the reconstructed image.
Recently, 
Katsevich et al   \cite{katsevich2012stability} proved  the general uniqueness results for the interior problem and provided stability estimates.
Using the total variation (TV) penalty, the authors in \cite{yu2009compressed} showed that a unique reconstruction is possible if the images are piecewise smooth.
In a series of papers \cite{ward2015interior,lee2015interior}, our group has shown that
a generalized L-spline along a collection of chord lines passing through the ROI can be uniquely recovered \cite{ward2015interior};
and we further substantiated that  the high frequency signal can be recovered analytically  thanks to the Bedrosian identify,  whereas the computationally expensive iterative reconstruction need only be performed to reconstruct the low frequency part of the signal after downsampling \cite{lee2015interior}.
While this approach significantly reduces the computational complexity of the interior reconstruction,
the computational complexity of existing iterative reconstruction algorithms  prohibits their  routine clinical use. 


%
%
%
%

In recent years, deep learning algorithms using convolutional neural network (CNN) 
have been successfully used for low-dose CT \cite{kang2017deep, chen2017low}, sparse view CT \cite{han2016deep,jin2017deep}, etc.
However, the more we have observed impressive empirical
results in CT problems, the more unanswered questions we encounter.
In particular, one of the most critical questions for biomedical applications is whether
a deep learning-based CT does create any artificial  structures that may mislead radiologists in their clinical decision.
Fortunately, 
in  a recent theory of {\em deep convolutional framelets} \cite{ye2017deep},
we showed that
the success of deep learning is not from a magical power of a black-box, but rather comes from the power of a novel signal representation using non-local basis combined with data-driven local basis.  
Thus, 
the deep network is indeed a natural extension of classical signal representation theory such as  wavelets, frames, etc; so rather than creating new informations,
it attempts to extract the
most information out of the the input data using the optimal signal representation.

Inspired these findings, here we propose a deep learning framework for interior tomography problem.
Specifically, 
we demonstrate that the interior tomography problem can be formulated as a reconstruction problem in an end-to-end manner
under the constraints that remove the null space signal components of the truncated Radon transform.
%
%
%
%
%
%
Numerical results 
confirmed the proposed deep learning architecture outperforms the existing interior tomography methods in image quality and reconstruction time.

\section{Theory}

\subsection{Problem Formulation}

Here, we consider 2-D interior tomography problem and follow the notation in \cite{ward2015interior}.
The variable $\thetab$ denotes a vector on the unit sphere $\Sd\in \Rd^2$.  The
collection of vectors that are orthogonal to $\thetab$ is denoted as
$$\thetab^\perp=\{\yb \in \Rd^2~:~\yb\cdot\thetab = 0 \}.$$
We refer to real-valued functions in the spatial domain
as images and denote them as $f(\xb)$ for $\xb \in  \Rd^2$.
We denote the Radon transform of an image $f$ as
\begin{eqnarray}
\Rc f(\thetab,s):= \int_{\thetab^\perp} f(s\thetab+\yb) d\yb
\end{eqnarray}
where $s\in \Rd$ and $\thetab \in \Sd$. The local
Radon transform for the truncated field-of-view is  the restriction of $\Rc f$ to the
region $\{(\thetab,s)~:~|s|<\mu \} ,$ which is denoted as
$\Tc _\mu \Rc f$.
Then, the  interior reconstruction is to find the unknown $f(\xb)$ within the ROI from
$\Tc_\mu\Rc f$.

\begin{figure}[!hbt]
\centering
\includegraphics[width=5cm]{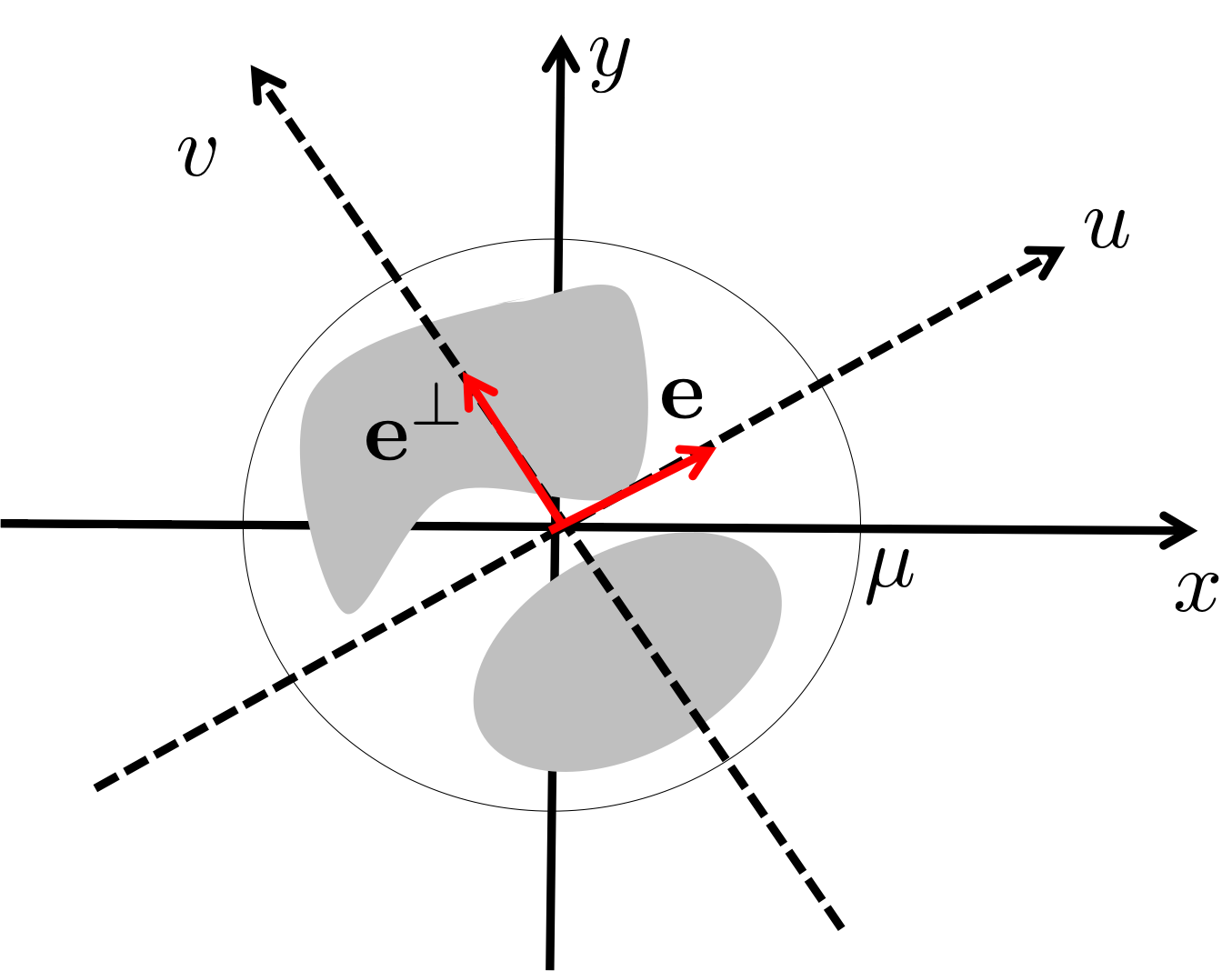} 
\caption{The coordinate system for interior tomography.}
\label{fig:coordinate}
\end{figure}

\subsection{Null Space of Truncated Radon Transform}

\begin{figure}[!bt]
\centering
\includegraphics[width=7cm]{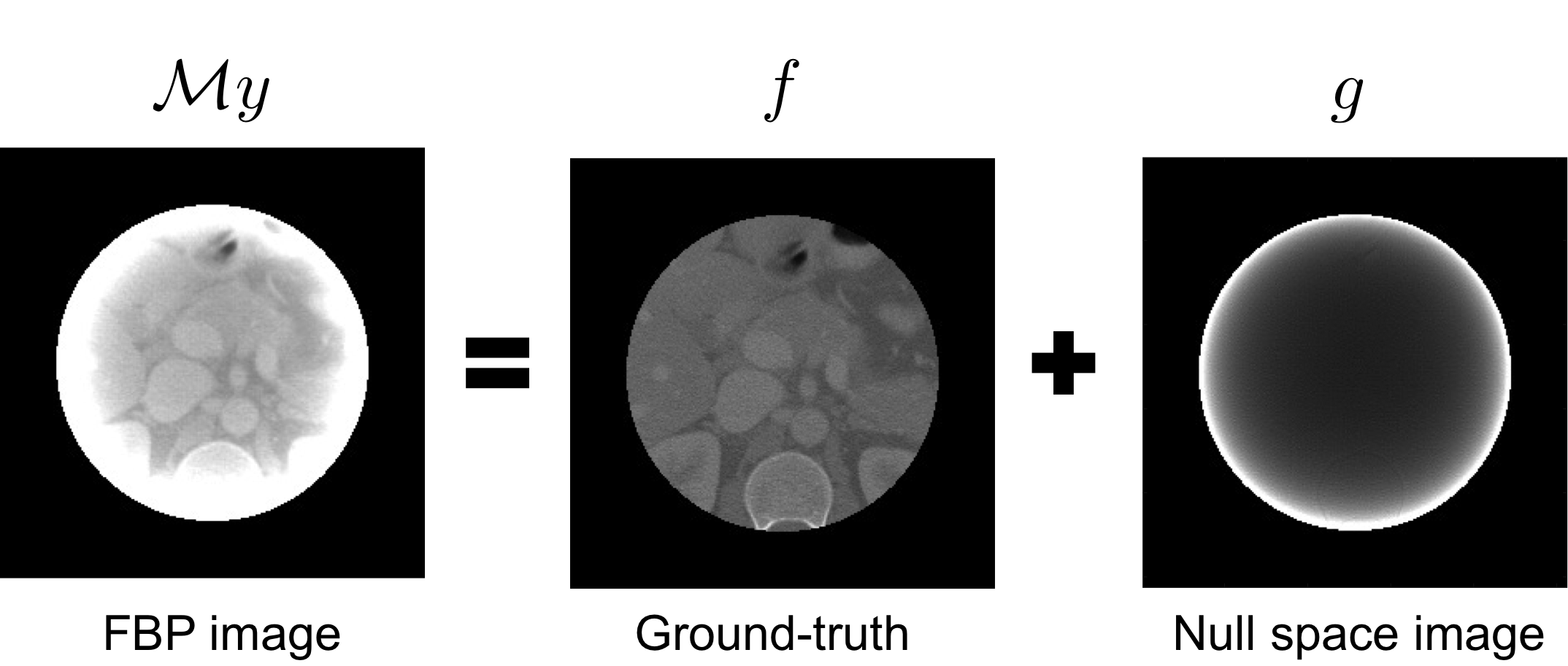} 
\caption{Decomposition of the analytic reconstruction into null space component
and the true image.}
\label{fig:null}
\end{figure}

The main technical difficulty of the interior reconstruction is
the existence of the null space \cite{ward2015interior, katsevich2012finite}. 
To analyze the null space, we follow the mathematical analysis in \cite{ward2015interior}.
Specifically,  the analytic inversion of $\Tc_\mu\Rc f$ can be equivalently represented
using the differentiated backprojection followed by the truncated Hilbert transform
along the chord lines,  so we analyze the interior reconstruction problem
to take advantages of this.
More specifically, if the unit vector  $\eb\in\Rd^2$
along the chord line   is set as a coordinate axis, then we can find the unit vector
 $\eb^\perp\in \Rd^2$ such that $V=[\eb,\eb^\perp]$ consists of the basis for the local coordinate system  and $(u,v) \in \Rd^2$ denotes its coordinate value (see
 Fig.~\ref{fig:coordinate}).
 We further define 1-D index set parameterized by the $v$:
 $$I_\mu(v) := \{u' \in \Rd ~|~\sqrt{(u')^2+v^2} \leq \mu \}.$$
Then,  the null space of the $\Tc_\mu \Rc f$ is given by \cite{ward2015interior,lee2015interior}:
%
%
%
%
%
\begin{equation}\label{eq:null}
\Nc_\mu:= \left\{ g ~|~ g(u,v)= -\int_{u'\notin I_\mu(v)} \frac{du'}{\pi(u-u')}\psi(u',v)\right\} \notag\\
\end{equation}
for some functions $\psi(u,v)$.
A typical example of the null space image $g$ is illustrated in Fig.~\ref{fig:null}.
This is often called as the cupping artifact. 
The cupping artifacts reduce contrast  and interfere with clinical diagnosis.

Note that the null space signal $g\in \Nc_\mu$  is differentiable in any order due to the removal of the origin in the integrand. Accordingly,  an interior
reconstruction algorithm needs an appropriate regularization term that suppresses $g\in \Nc_\mu$ by exploiting this.
Specifically,  one could find an analysis transform 
$\mathrm{L}$ such that its null space  $\Nc_{\mathrm{L}}$ is composed of entire function,
and use it for an analysis-based regularization term.
For example, the regularization using TV \cite{yu2009compressed}  and L-spline model \cite{ward2015interior,lee2015interior} correspond to this.
The main result on the perfect reconstruction in  \cite{ward2015interior} is then stated as follows.  If the null space component $g \in \Nc_\mu$ is equivalent
to a signal $h\in \Nc_{\mathrm{L}}$ within the ROI, then $g$ is identically zero due to the
characterization of Hilbert transform pairs as boundary
values of analytic functions on the upper half of the complex plane \cite{ward2015interior};
so TV or L-spline regularization provides the unique solution.

\subsection{CNN-based Null Space Removal}

Instead of designing a linear operator ${\mathrm{L}}$ such that  the common null space of $\Nc_\mu$ and $\Nc_{\mathrm{L}}$ to be zero,
we can design a frame $\Wc$ and its dual $\tilde \Wc$ such that
$\tilde \Wc^\top \Wc=I$ and
$\tilde \Wc^\top S_\lambda \Wc (f^*+g)  = f^*$ for all 
$g \in \Nc_\mu$ and  the ground-truth image $f^*$.
This frame-based regularization is also an active field of research for image denoising,
inpainting, etc \cite{cai2008framelet}.

 
One of the most important contributions of the deep convolutional framelet theory \cite{ye2017deep}
is that $\Wc$ and $\tilde \Wc^\top$ correspond to the encoder and decoder structure of a CNN, respectively,
and the shrinkage operator $S_\lambda$ emerges by controlling the number of filter channels and nonlinearities.
Accordingly,  a convolutional neural network    represented by 
$\Qc =\tilde \Wc^\top S_\lambda \Wc$ can be  designed   such that
\begin{eqnarray}\label{eq:Qc2}
\Qc(f^*+g) = f^* ,\quad \forall g \in \Nc_\mu.
\end{eqnarray}
%
Then,  our interior tomography algorithm is formulated to find the solution $f$ for the following 
problem:
\begin{eqnarray}\label{eq:constraint}
y  = \Tc_\mu\Rc f ,\quad  \Qc f = f^* \  ,
\end{eqnarray}
where $f^*$ denotes the ground-truth data available for training data, and $\Qc$ denotes the
CNN satisfying \eqref{eq:Qc2}.
Now, by defining $\Mc$ as a right-inverse of $\Tc_\mu \Rc$, i.e. $(\Tc_\mu\Rc)\Mc y =y, \forall y$,
we have
$$\Mc y = f^*+g$$
for some $g\in \Nc_\mu$, since the right inverse is not unique due to the existence of the null space.
See Fig.~\ref{fig:null} for the decomposition of $\Mc y$. Thus, $\Mc y$ is a feasible solution for \eqref{eq:constraint},
since 
\begin{eqnarray}\label{eq:Q2}
\Qc \Mc y = \Qc \left(f^*+g\right) = f^*,
\end{eqnarray}
and the data fidelity constraint is automatically satisfied due to the definition of the right inverse.
Therefore, the neural network training problem to satisfy \eqref{eq:Q2} can be equivalently
represented by
\begin{eqnarray}\label{eq:opt}
\min_{\Qc} \sum_{i=1}^N\|f_i^* - \Qc \Mc y_i \|^2
\end{eqnarray}
where $\{(f_i^*,y_i)\}_{i=1}^N$ denotes the training data set composed of ground-truth image an its truncated projection.
A typical example of the right inverse for the truncated Radon transform is the inverse Radon transform, which
can be implemented by the filtered backprojection (FBP) algorithm.
Thus, $\Mc y_i$ in \eqref{eq:opt} can be implemented using the FBP.

After the neural network $\Qc$ is learned, the inference can be done simply 
by processing  FBP  reconstruction image from a truncated radon data $y_t$ using the neural network
$\Qc$, i.e. $\hat f =\Qc\Mc y_t$. 
The details of the network $\Qc$ and the training procedure will be discussed in the following section.

%
%
%


%
%
%

\section{Method}

\subsection{Data Set}


Ten subject data sets 
from AAPM Low-Dose CT Grand Challenge were used in this paper.
Out of ten sets, eight sets were used for network training.
The other two sets were used for validation and test, respectively.
The provided data sets were originally acquired in helical CT, and were rebinned from the helical CT to $360^{\circ}$ angular scan fan-beam CT.
The $512 \times 512$ size artifact free CT images are reconstructed from the rebinned fan-beam CT data using FBP algorithm.
From the CT image, sinogram is numerically obtained using a forward projection operator.
The number of detector in numerical experiment is 736. 
Only 350 detectors in the middle of 736 detectors are used to simulate the truncated projection data.
Using this, we reconstruct $256 \times 256$ ROI images.

\subsection{Network Architecture}

\begin{figure}[t]
\centering
\includegraphics[width=8.5cm]{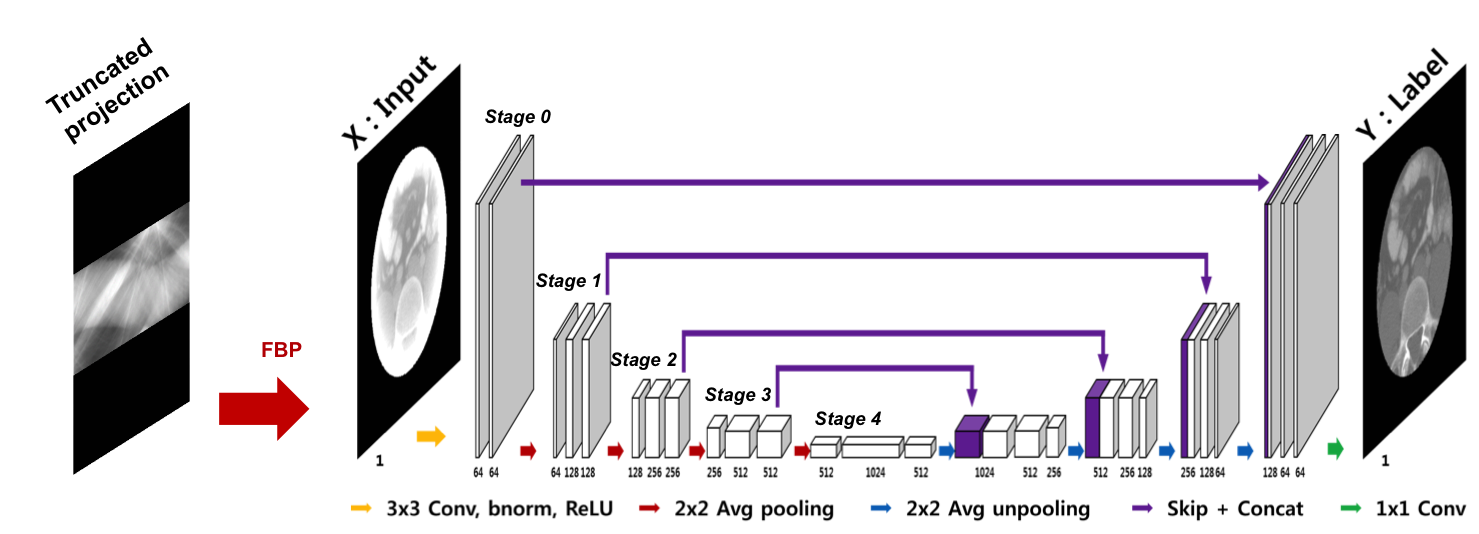} 
\caption{The proposed deep learning architecture for inteior tomography.}
\label{fig:architecture}
\end{figure}

The proposed network is shown in Fig. \ref{fig:architecture}.
The first layer is the FBP layer that reconstructs the cupping-artifact corrupted images from the truncated projection data, which is followed by
a modified architecture of U-Net \cite{ronneberger2015u}.
A yellow arrow in Fig. \ref{fig:architecture} is the basic operator and consists of $3 \times 3$ convolutions followed by a rectified linear unit and batch normalization.
The yellow arrows between the seperate blocks at every stage are omitted.
A red arrow is a $2 \times 2$ average pooling operator and is located between the stages.
The average pooling operator doubles the number of channels and reduces the size of the layers by four.
Conversely, a blue arrow  is $2 \times 2$ average unpooling operator, reducing the number of channels by half and increasing the size of the layer by four. 
A violet arrow is the skip and concatenation operator. 
A green arrow is the simple $1 \times 1$ convolution operator generating the final reconstruction image.

\subsection{Network training}

The proposed network was implemented using MatConvNet toolbox in MATLAB R2015a environment. 
Processing units used in this research are Intel Core i7-7700 central processing unit and GTX 1080-Ti graphics processing unit.
Stochastic gradient reduction was used to train the network. 
As shown in Fig.~\ref{fig:architecture}, the inputs of the network are  the truncated projection data, i.e. $y_i$.
The target data $f_i$  corresponds to the  256 $\times$ 256 size center ROI image cropped from the ground-truth data.
The number of epochs was 300. 
The initial learning rate was $10^{-3}$, which gradually dropped to $10^{-5}$. 
The regularization parameter was $10^{-4}$.
Training time lasted about 24 hours.

\section{Results}

\begin{figure}[t]
\centering
\includegraphics[width=8.cm]{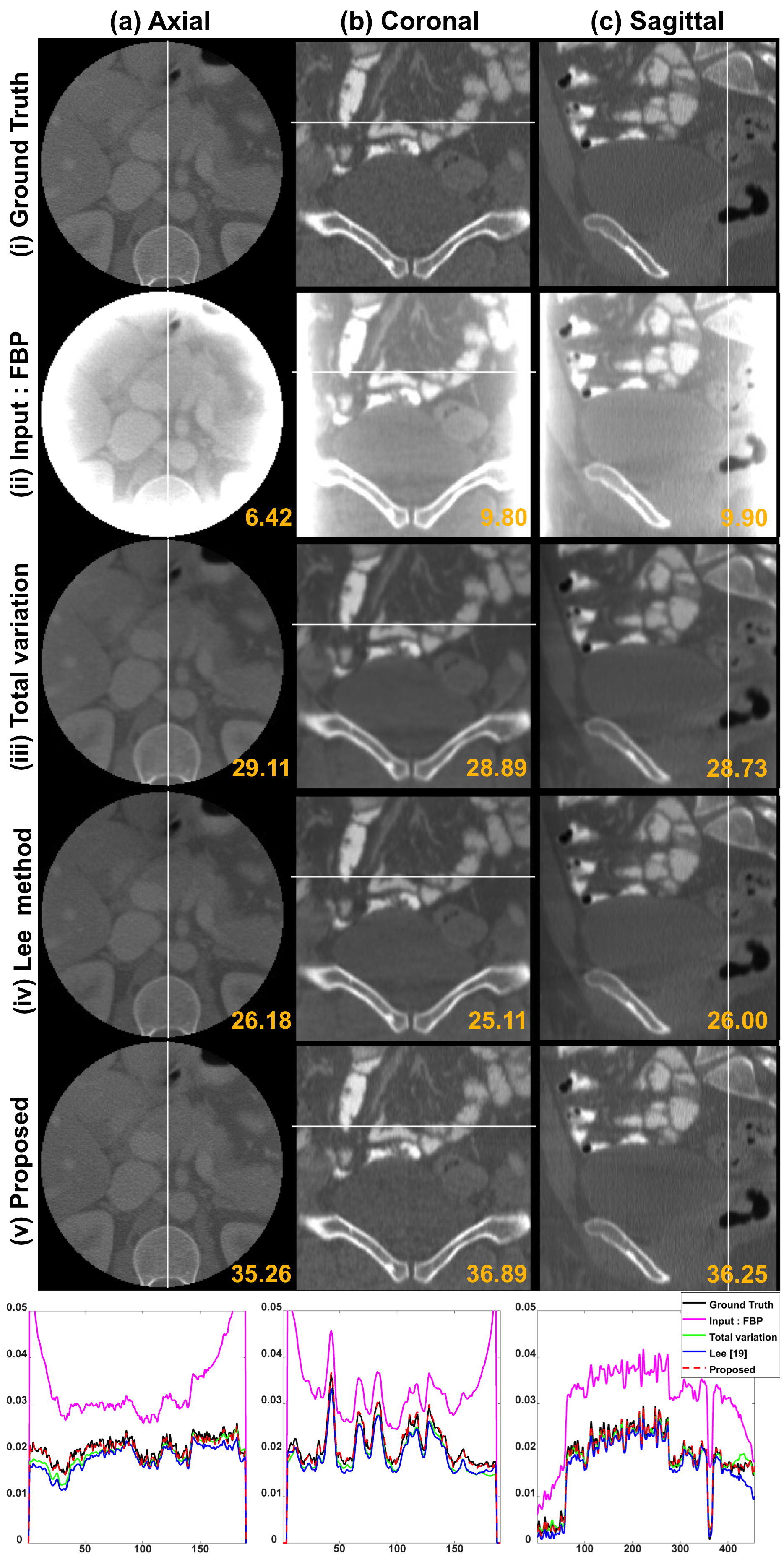} 
\caption{Reconstruction images by the cone-beam simulation. The last row shows the cut-view plots of the white lines on the images. 
The number written in the images is the PSNR value in dB.}
\label{recon_results}
\end{figure}

\begin{figure}[t]
\centering
\includegraphics[width=7.cm]{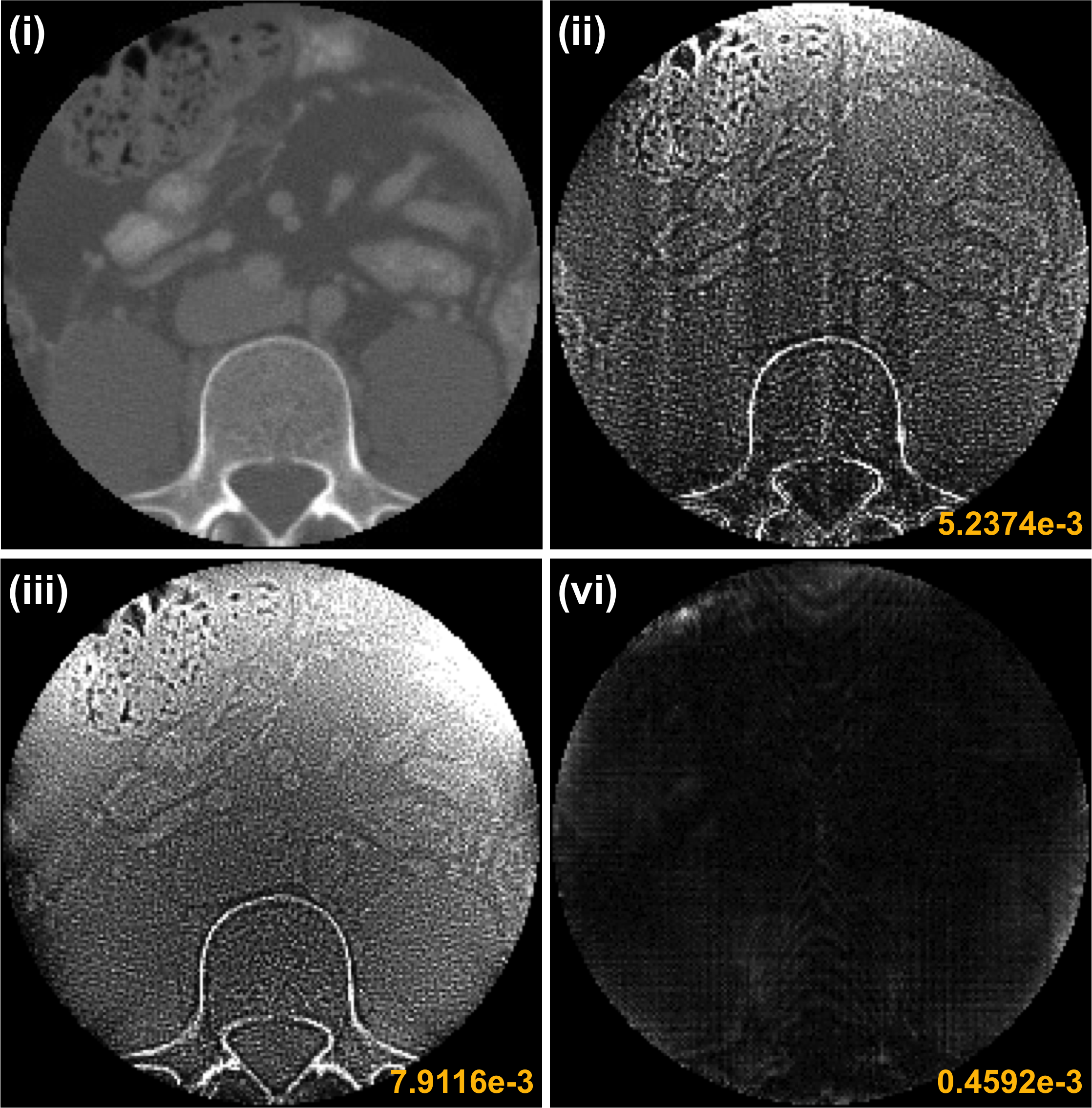} 
\caption{Error images from (ii) TV, (iii) Lee method \cite{lee2015interior}, and (iv) the proposed method. (i) is ground truth image. The number written in the images is the NMSE value.}
\label{difference}
\end{figure}

We compared the proposed method with existing iterative methods such as
the TV penalized reconstruction \cite{yu2009compressed} and the L-spline based multi-scale regularization method by Lee et al \cite{lee2015interior}.
Fig. \ref{recon_results} shows the ground-truth images and reconstruction results by 
FBP, TV,  Lee method \cite{lee2015interior} and the proposed method.
The graphs in the bottom row in Fig.  \ref{recon_results} are the cross-section view along the white lines on the each images.
Fig. \ref{difference} shows the magnitude of difference images between the ground truth image and reconstruction results of each method.
The reconstructed images and the cut-view graphs in Fig. \ref{recon_results} show that the proposed method results have more fine details than the other methods.
The error images in Fig. \ref{difference} confirm that the high frequency components such as edges and textures are better restored in the proposed method than other method.

We also calculated the average values of the peak signal-to-noise ratio (PSNR) and the normalized mean square error (NMSE) in Table ~\ref{tab_quality}.
The proposed method achieved the highest value in PSNR and the lowest value in NMSE with about $7\sim 10$ dB improvement.
The computational times for TV, Lee method \cite{lee2015interior} and the
proposed method were 1.8272s,
0.3438s, and
0.0532s, respectively, for each slice reconstruction.
The processing speed of the proposed method is about 34 times faster than the TV method and 6 times faster than  Lee  method \cite{lee2015interior}.

\begin{table}[h!] 
\caption{Quantitative comparison of various  methods.}
\vspace*{-0.5cm}
\label{tab_quality}
\begin{center}
\begin{tabular}{c|cccc}
\hline
&  FBP &  TV  & Lee method \cite{lee2015interior} &  Proposed\\
\hline\hline
PSNR [dB] &
9.4099 &
30.2004 &
27.0344 &
37.4600 \\
NMSE &
8.2941e-1 &
6.9137e-3 &
1.4332e-2 &
1.2994e-3 \\
\hline
\end{tabular}
\end{center}
\end{table}

%

\section{Conclusion}

In this paper, we proposed a deep learning network for interior tomography problem.
The reconstruction problem was formulated as a constraint  optimization problem under
data fidelity and null space constraints.
Based on the theory of deep convolutional framelet,
the null space constraint was implemented using the convolutional neural network with encoder and decoder architecture.
Numerical results showed that the proposed method has the highest value in PSNR and the lowest value in NMSE and the fastest computational time.



\section*{Acknowledgment}
The authors would like to thanks Dr. Cynthia McCollough,  the Mayo Clinic, the American Association of Physicists in Medicine (AAPM), and grant EB01705 and EB01785 from the National Institute of Biomedical Imaging and Bioengineering for providing the Low-Dose CT Grand Challenge data set.
This work is supported by National Research Foundation of Korea, Grant number NRF-2016R1A2B3008104. 
Yoseob Han and Jawook Gu contributed equally to this work.

 

%





\end{document}